\documentclass[10pt,twocolumn,letterpaper]{article}
\def\confName{ICCV}
\usepackage{iccv}              
\usepackage{multirow}

\definecolor{iccvblue}{rgb}{0.21,0.49,0.74}
\usepackage[pagebackref,breaklinks,colorlinks,allcolors=iccvblue]{hyperref}

\title{RAUM-Net : Regional Attention and Uncertainty-aware Mamba Network}

\author{Mingquan Liu\\
China University of Geosciences\\
388 Lumo Road, Wuhan, 430074, Hubei, China\\
{\tt\small wxqnl@cug.edu.cn}
}

\begin{document}
\maketitle
\begin{abstract}
Fine-Grained Visual Categorization (FGVC) remains a challenging task in computer vision due to subtle inter-class differences, fragile feature representations, and vulnerability to occlusion. Although classification on natural image datasets has made significant progress over the years, existing methods still struggle in fine-grained scenarios—especially in real-world applications where unlabeled data is abundant but labeled data is scarce. Under such conditions, it is difficult to balance accuracy and robustness. Furthermore, since partial occlusion is common in practical datasets, effectively learning fine-grained feature differences from limited data is key to improving both detection performance and practical usability.To address these challenges, we propose a semi-supervised fine-grained image classification method that combines Mamba-based feature modeling, region attention, and a Bayesian uncertainty module. The Mamba module enhances local-to-global feature modeling, while the region attention mechanism strengthens the model’s focus on key areas during semi-supervised learning. In addition, a Bayesian inference strategy is employed to select high-quality pseudo-labels, improving the stability and reliability of the training process. Experimental results show that the proposed method achieves strong performance on multiple FGVC benchmark datasets with randomly introduced occlusions. In particular, it demonstrates better robustness and higher classification accuracy under severe occlusion or when labeled data is extremely limited. These results indicate the method's potential in addressing practical and challenging semi-supervised FGVC tasks. 
The code for this method is available at: \href{https://github.com/wxqnl/RAUM-Net}{https://github.com/wxqnl/RAUM-Net}.
\end{abstract}

\section{Introduction}
Fine-Grained Visual Classification (FGVC) presents a significant challenge in computer vision, focusing on distinguishing visual object categories where inter-class differences are minute, yet intra-class variations in morphology or appearance are substantial. Typical examples include differentiating bird species, identifying specific car models, or recognizing particular plant varieties \cite{huang2016part,wang2018learning}. Accurate FGVC capabilities are crucial for a growing range of real-world applications requiring high precision, such as automated biodiversity monitoring, sophisticated industrial quality control, and advanced intelligent security systems \cite{kong2017low}. However, the inherent difficulty of fine-grained recognition lies in the fact that discriminative features are often subtle and localized, making them highly susceptible to environmental disturbances like occlusion, viewpoint changes, and illumination fluctuations. Consequently, traditional image classification techniques often perform inadequately on such tasks \cite{CHEN2024110265}.

In recent years, advancements in deep neural networks, particularly Convolutional Neural Networks (CNNs) and Vision Transformers, have led to considerable progress in feature representation and classification performance for FGVC. Nevertheless, the effectiveness of these methods typically relies heavily on extensive, precisely labeled data for fully supervised training. Acquiring such large-scale, high-fidelity fine-grained datasets is not only costly and time-consuming but also prone to introducing subjective annotation biases, significantly limiting further improvements in FGVC model performance and their widespread deployment in practical scenarios.

To mitigate the strong reliance on abundant labeled data, Semi-Supervised Learning (SSL) \cite{ouali2020overview} has emerged as a compelling paradigm in the FGVC domain. SSL methods strategically leverage a combination of ample unlabeled data and limited labeled data to enhance model generalization and substantially reduce annotation costs. While mainstream SSL strategies for image classification, such as pseudo-label generation \cite{xu2021end} and consistency regularization \cite{sohn2020fixmatch}, have achieved success in general image classification tasks, their direct application to the demanding FGVC context encounters two primary challenges.

Firstly, the unreliability of pseudo-labels is significantly amplified. Given the inherently ambiguous boundaries between fine-grained categories, models are more prone to errors when generating predictions on unlabeled data. Even minor prediction inaccuracies can lead to incorrect pseudo-label assignments, potentially causing error accumulation during training and, in severe cases, misguiding model optimization, thereby degrading final performance. This issue of pseudo-label noise is particularly pronounced in FGVC compared to general classification tasks.

Secondly, the prevalent occlusion issue in real-world scenarios is often overlooked by existing SSL methods. In practical FGVC applications, target objects, such as birds partially obscured by foliage, are frequently observed in an incomplete state. Occlusion not only directly masks key discriminative fine-grained features, increasing the difficulty of feature extraction, but more importantly, it significantly reduces model confidence in predictions for unlabeled samples. This further exacerbates the negative impact of inaccurate pseudo-labels, limiting the effectiveness of semi-supervised training.

Addressing these intertwined challenges of difficult feature discrimination under occlusion and susceptibility of pseudo-labels to noise in semi-supervised FGVC, this study proposes a novel semi-supervised fine-grained image classification method. This approach integrates the Mamba architecture \cite{gu2023mamba} with a Region Attention and Bayesian Uncertainty (RABU) module. Mamba, as an advanced state-space model, excels at capturing long-range dependencies while demonstrating robust local fine-grained feature modeling capabilities. Its inherent structural properties make it promising for extracting more robust and discriminative features when dealing with partially occluded fine-grained objects, thus making it our chosen core feature extractor. Building upon this, we incorporate the RABU module during the semi-supervised learning process to collaboratively tackle the core challenges of feature ambiguity and pseudo-label noise. Specifically, the RABU module comprises two key components. The region attention component dynamically guides the model to focus on the most informative and unobscured discriminative regions within an image. By enhancing attention to critical visual cues, it effectively suppresses interference from occluders and background noise on feature representations, thereby improving the robustness of feature representations in complex environments. The Bayesian uncertainty component, on the other hand, introduces Bayesian inference principles \cite{gal2016dropout} to quantify model confidence in predictions for unlabeled samples. Based on this quantified uncertainty, a principled pseudo-label selection mechanism is constructed, involving only high-confidence pseudo-labels in model training. This approach significantly alleviates the negative impact of pseudo-label noise on the semi-supervised learning process, enhancing training stability and the reliability of the final model.The main contributions of this paper can be summarized as follows:
\begin{itemize}
    \item We propose RAUMNet, a novel semi-supervised FGVC framework that integrates the Mamba architecture with an innovative Region Attention and Bayesian Uncertainty module. This framework leverages Mamba for deep feature capture and the RABU module to collaboratively address feature ambiguity under occlusion and pseudo-label noise in semi-supervised learning.
    \item Within the RABU module, we design and implement an organic combination of a region attention mechanism and a pseudo-label selection strategy based on Bayesian uncertainty. The region attention component enhances the model's discriminative feature extraction capabilities in occluded scenarios, while the uncertainty selection mechanism improves the robustness and efficiency of semi-supervised training using unlabeled data.
    \item Comprehensive and systematic experimental validation is conducted on multiple standard fine-grained classification benchmark datasets, including their variants with varying degrees of occlusion. The experimental results strongly demonstrate that RAUMNet outperforms existing state-of-the-art semi-supervised classification methods across different occlusion conditions, showcasing its excellent adaptability and robustness.
\end{itemize}

\section{Related Work}

\subsection{Semi-supervised Learning for Fine-Grained Visual Classification}
Fine-Grained Visual Classification (FGVC), due to its focus on subtle differences between subcategories, typically necessitates a substantial amount of precisely annotated data to achieve optimal performance \cite{9295723}. However, the high cost of fine-grained annotation has motivated researchers to explore Semi-Supervised Learning (SSL) to leverage abundant unlabeled data. Early SSL explorations, exemplified by Pseudo-Label, involved supervised learning by generating pseudo-labels for unlabeled data. Subsequently, methods based on consistency regularization, such as Mean Teacher, and strategies combining pseudo-labels have continuously evolved. Among these, FixMatch \cite{sohn2020fixmatch} and its variants, FlexMatch \cite{zhang2021flexmatch} and FreeMatch \cite{wang2022freematch}, which generate pseudo-labels via weak augmentation and enforce consistency constraints with strong augmentation while incorporating a confidence threshold for selection, have become mainstream paradigms in SSL, significantly improving performance.

Nevertheless, when directly applying general SSL methods to FGVC, the high similarity between fine-grained categories leads to more confused model predictions, causing a drastic amplification of the pseudo-label noise problem. Although some works \cite{mugnai2022fine} have attempted to integrate domain knowledge or specific learning strategies into SSL-FGVC, most methods still struggle to fully overcome the unreliability of pseudo-labels stemming from subtle feature differences and potential interference. This study aims to address the core challenges in SSL-FGVC by employing more powerful feature representations and a more robust pseudo-label processing mechanism.

\subsection{Occlusion in Fine-Grained Visual Classification}
Occlusion is a critical factor affecting the practical performance of FGVC as it frequently obscures local details essential for distinguishing subcategories. Early research attempted to locate and utilize visible key parts through Part-based Models. With the advancement of deep learning, attention mechanisms have been widely adopted to enable models to automatically focus on discriminative regions within an image, thereby indirectly enhancing tolerance to partial occlusion. Examples include the iterative attention in RA-CNN, salient region navigation in NTS-Net \cite{yang2018learning}, and Transformer-based methods like TransFG that utilize self-attention to capture global and local relationships.

Further research has explored more direct strategies for handling occlusion, such as feature reconstruction and adversarial erasure \cite{stammes2021find,yang2020dfr}, aiming to learn features invariant to occlusion or capable of inferring occluded content. Learning Counterfactual Representations attempts to disentangle inherent object features from confounding factors like occlusion or background. On the other hand, generative models, particularly Diffusion Models, are being explored for feature repair or generating more robust representations. One study \cite{ho2020denoising} utilized diffusion models for prototype learning, indirectly improving robustness against interference. While significant research has addressed occlusion in supervised learning, the issues introduced by occlusion in semi-supervised learning frameworks are more complex: it not only disrupts feature extraction but also severely reduces model prediction confidence for unlabeled samples, thereby greatly diminishing the effectiveness of SSL methods relying on pseudo-labels. Effective solutions for occlusion in semi-supervised FGVC remain insufficient. The region attention mechanism introduced by RAUMNet is precisely designed to explicitly guide the model to focus on unobscured regions within the SSL process, aiming to learn reliable fine-grained features even in the presence of occlusion.

\subsection{Visual Feature Extraction Architectures}
Powerful feature representation is fundamental to solving complex visual tasks. Convolutional Neural Networks (CNNs) \cite{yu20251d} have achieved immense success through hierarchical extraction of local features, but their inherent local receptive fields limit effective modeling of global context. Vision Transformer (ViT) architectures \cite{dosovitskiy2020image} and their variants, such as Swin Transformer \cite{liu2022swin}, have drawn inspiration from NLP successes, utilizing self-attention mechanisms for global interaction and setting new records on various visual benchmarks. However, they also introduce significant computational complexity.

In recent years, the research community has been actively exploring novel efficient architectures beyond CNNs and ViTs. State Space Models (SSMs), with their potential for modeling long-range dependencies at linear complexity, have garnered widespread attention. Mamba, as a representative work in this area, has demonstrated excellent performance and efficiency in sequence modeling tasks through its selective scan mechanism and hardware-aware design. Consequently, researchers have begun applying it to the visual domain. Early in 2024, a series of visual Mamba variants emerged, such as Vim \cite{wu2025dynamic} and VMamba \cite{liu2024vmamba}, exploring how to effectively integrate SSMs into 2D visual feature extraction, positioning them as potential next-generation visual backbone networks.

Concurrently, attention mechanisms, as effective means of enhancing feature representation and focusing on critical information, whether classic channel and spatial attention or more complex mechanisms aimed at locating discriminative regions, continue to play a crucial role in FGVC. The design of RAUMNet is based on considerations of these advancements, prospectively adopting Mamba as the foundational architecture to leverage its efficient global context modeling capabilities. This is complemented by a customized region attention mechanism to strengthen the capture of local details vital for FGVC tasks, forming a complementary advantage.

\begin{figure*}[!ht]
\centering
\includegraphics[width=2\columnwidth]{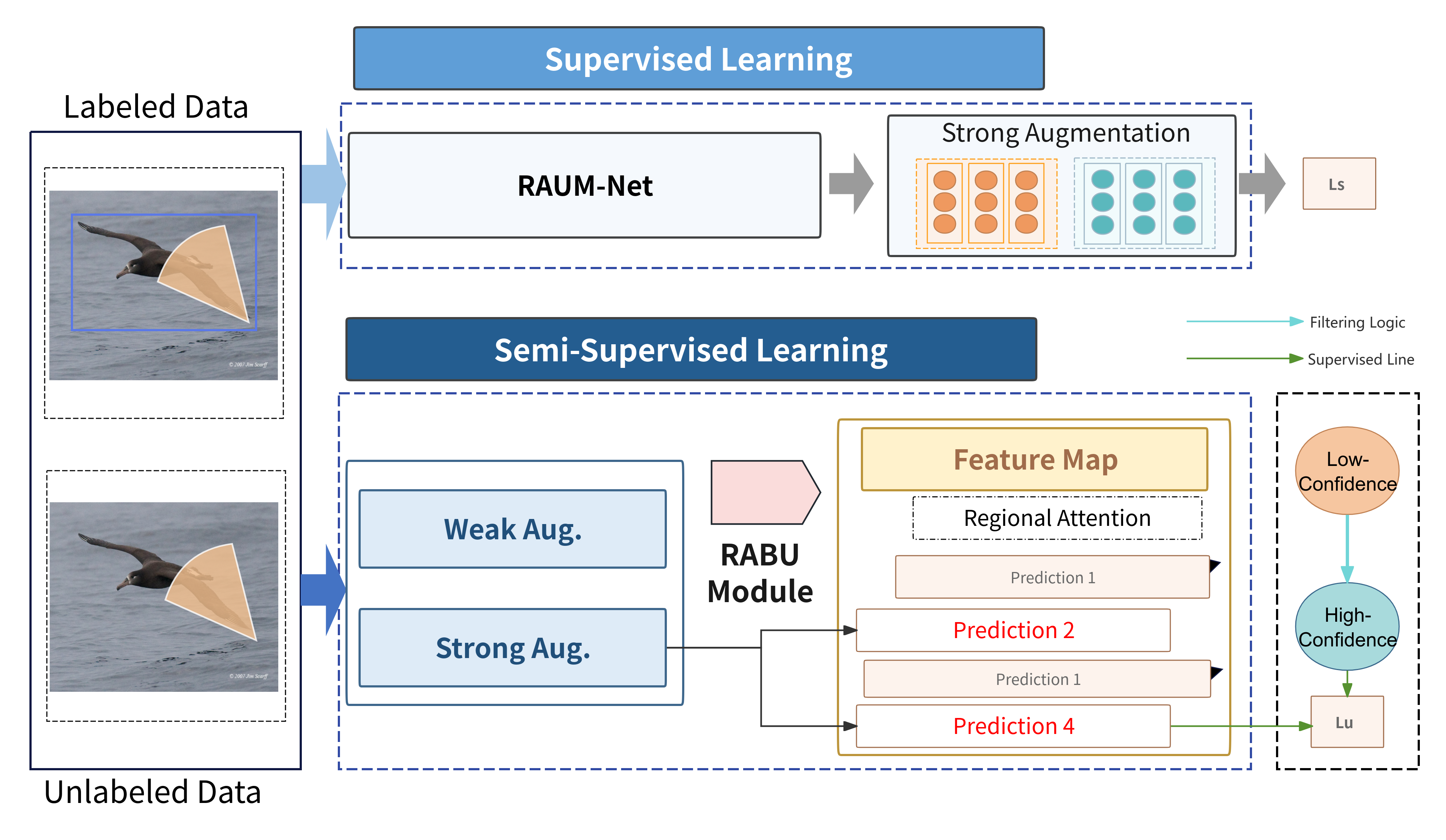}
    \caption{The overall framework of RAUM-Net. The supervised learning path utilizes labeled data for standard cross-entropy loss, while the semi-supervised learning path employs weak and strong augmentation to generate pseudo-labels and enforce consistency, with the RABU module filtering low-confidence predictions.}
\label{fig:framework}
\end{figure*}

\subsection{Uncertainty Awareness in Semi-supervised Learning}
The quality of pseudo-labels is critical to the success of SSL. In FGVC tasks with low class distinctiveness, models are more prone to generating uncertain or erroneous predictions, making the pseudo-label noise problem particularly prominent. FixMatch simply relies on the maximum value of model outputs as confidence and sets a fixed threshold, which has proven insufficient in complex scenarios. Therefore, quantifying and utilizing model prediction uncertainty has become an important avenue for improving SSL reliability.

Early methods such as entropy minimization indirectly encouraged models to produce more confident predictions, while Bayesian Neural Networks (BNNs) provide a theoretical framework for modeling uncertainty but are typically computationally complex. Monte Carlo Dropout (MC Dropout), as a practical and easily implementable approximate Bayesian inference method, is widely used in deep learning for estimating model uncertainty. Building on these ideas, a series of uncertainty-aware SSL methods have been proposed. UPSR-SSL utilizes estimated uncertainty to filter low-confidence pseudo-labels or to weight pseudo-label losses \cite{zhang2025uncertainty}. Alternatively, UPS's curriculum learning philosophy \cite{rizve2021defense} is employed. The most recent research, AnomalyMatch \cite{gomez2025anomalymatch}, even proposes directly correcting model confidence outputs to better align with true accuracy. The core trend in this field is shifting from simply rejecting low-confidence samples to a more nuanced utilization of uncertainty information to guide the entire learning process.

The Bayesian uncertainty component in RAUMNet draws inspiration from advancements in this research direction. It employs uncertainty quantification based on Bayesian inference and utilizes it to construct a principled pseudo-label selection mechanism. This enables the model to intelligently identify and discard high-risk pseudo-labels, prioritizing learning from knowledge that the model truly masters. Consequently, in the challenging domain of semi-supervised FGVC, it significantly enhances training stability and the final model's generalization capability.

\section{Method}
To address the severe challenges in Semi-Supervised Fine-Grained Visual Classification (FGVC) arising from subtle class differences, fragile feature representations, and pervasive occlusion interference, we propose a novel semi-supervised learning framework named RAUM-Net (Regional Attention and Uncertainty-aware Mamba Network). The overall architecture is illustrated in Figure \ref{fig:framework}. RAUM-Net utilizes the robust Mamba architecture as its core feature extraction backbone, aiming to leverage its efficient long-range dependency modeling capabilities to capture fine-grained discriminative features. Building upon this, we design and integrate an innovative Regional Attention and Bayesian Uncertainty (RABU) module to collaboratively enhance semi-supervised learning performance in complex environments. Specifically, to bolster feature robustness under occlusion, the regional attention component within the RABU module is employed to dynamically guide the model to focus on key, unobscured regions of an image. Concurrently, for unlabeled data, the Bayesian uncertainty component effectively filters out low-quality pseudo-labels caused by occlusion or class confusion through a principled pseudo-label selection strategy, thereby ensuring the stability and reliability of semi-supervised training. The structural design of RAUM-Net comprehensively considers the subtle and easily disturbed nature of discriminative features in FGVC tasks, aiming to significantly improve classification accuracy and robustness under label scarcity and occlusion conditions.

\subsection{Overall Framework}

The semi-supervised learning process of RAUM-Net is built upon the successful principles of pseudo-labeling and consistency regularization. Given a training set containing a small amount of labeled data $D_L = \{(x_i^l, y_i^l)\}_{i=1}^{N_L}$ and a large quantity of unlabeled data $D_U = \{x_j^u\}_{j=1}^{N_U}$, our framework processes labeled and unlabeled samples in parallel during each training iteration.

As illustrated in Figure \ref{fig:framework}, for a mini-batch composed of $B_L$ labeled samples and $B_U$ unlabeled samples, the learning process proceeds as follows:

\begin{enumerate}
    \item \textbf{Supervised Path}: For each labeled sample $x_i^l$, we apply strong data augmentation $A_{strong}(\cdot)$, then input it into the RAUM-Net model $f_\theta$ for prediction, and compute the standard cross-entropy loss against the ground truth label $y_i^l$.

    \item \textbf{Semi-supervised Path}: For each unlabeled sample $x_j^u$, we adopt a two-stream strategy:
    \begin{itemize}
        \item \textit{Pseudo-Label Generation Stream}: The sample $x_j^u$ is subjected to weak data augmentation $A_{weak}(\cdot)$ (e.g., random flip and crop only), and then fed into the model $f_\theta$. During this process, the RABU module is activated to generate a high-quality pseudo-label $\hat{y}_j^u$. This process involves regional attention and uncertainty evaluation, with only rigorously screened samples producing effective pseudo-labels.
        \item \textit{Consistency Learning Stream}: The same unlabeled sample $x_j^u$ is subjected to strong data augmentation $A_{strong}(\cdot)$, and then input into the model $f_\theta$ to obtain the prediction probability distribution $p_j^u$.
    \end{itemize}

    \item \textbf{Loss Computation}: If the sample $x_j^u$ successfully generates a pseudo-label in the generation stream, this pseudo-label $\hat{y}_j^u$ is used to supervise the output $p_j^u$ from the consistency learning stream, and their cross-entropy loss is computed. The final total loss is constituted by the weighted sum of the supervised loss and the effective unsupervised consistency loss.
\end{enumerate}

\subsection{Mamba Backbone for Feature Extraction}
The core challenge in fine-grained visual classification lies in capturing subtle features critical for distinguishing subcategories, which may be distributed across different image locations. Traditional CNNs are limited by their local receptive fields, while ViTs, although capable of capturing global relationships, incur high computational costs. We select the Mamba architecture as the backbone of RAUM-Net due to its ability to efficiently model long-range dependencies at linear complexity, and its selective scan mechanism allows it to dynamically focus on information-rich local regions based on input content.

In our implementation, the input image $x \in \mathbb{R}^{H \times W \times 3}$ is first partitioned into a sequence of non-overlapping image patches. These patches are then converted into a 1D token sequence via a linear embedding layer. This sequence is subsequently fed into a deep network composed of stacked Mamba blocks. Each Mamba block efficiently models the sequence context through its core Selective State Space Model (S2M) mechanism. Finally, the Mamba backbone outputs a 2D feature map $F \in \mathbb{R}^{H' \times W' \times C}$, where $H'$ and $W'$ are the spatial dimensions of the feature map, and $C$ is the number of channels. This feature map $F$ provides rich feature representations, encompassing both global and local information, for the subsequent RABU module.

\subsection{Regional Attention and Bayesian Uncertainty Module}
The RABU module is the core of our method and is seamlessly integrated into the pseudo-label generation stream. It is designed to address two intertwined challenges: feature ambiguity caused by occlusion and pseudo-label noise in semi-supervised learning. This module consists of two components that work collaboratively.

\subsubsection{Regional Attention Component}
In occluded scenarios, irrelevant background or occluders introduce noisy features, significantly hindering the model's ability to recognize the main object. The purpose of the RA component is to proactively and explicitly guide the model to focus on the most discriminative and unobscured regions of the image before the final prediction.

Specifically, given the feature map $F$ output by the Mamba backbone, we design a lightweight attention generation network $g_{att}$ (implemented with two convolutional layers) to produce a spatial attention map $A \in \mathbb{R}^{H' \times W'}$. The computation process is as follows:
$$
A = \sigma(\text{Conv}_{3\times3}(\text{ReLU}(\text{Conv}_{1\times1}(F))))
$$
Here, $\text{Conv}_{1\times1}$ and $\text{Conv}_{3\times3}$ represent 1x1 and 3x3 convolutional operations, respectively. ReLU is the non-linear activation function, and $\sigma$ is the Sigmoid function, which normalizes the output attention values to the range $[0, 1]$, representing the importance of each spatial location.

Subsequently, we multiply this attention map $A$ with the original feature map $F$ element-wise (i.e., Hadamard product) to obtain the attention-weighted feature map $F_{att}$:
$$
F_{att} = F \odot A
$$
In this manner, the features in $F_{att}$ are re-weighted, causing the subsequent classifier to rely more heavily on regions assigned high attention weights (i.e., key discriminative regions), while features from occluded or background regions are effectively suppressed.

\subsubsection{Bayesian Uncertainty Component}
Simply relying on the maximum value of the Softmax output as confidence to filter pseudo-labels is unreliable in FGVC tasks where classes are highly similar. We introduce the concept of Bayesian inference to more robustly quantify the model's prediction uncertainty. We employ Monte Carlo Dropout (MC Dropout) as a practical and efficient approximation for Bayesian inference.

When generating pseudo-labels for weakly augmented unlabeled samples $A_{weak}(x_j^u)$, we keep the Dropout layers in the model active and perform $T$ random forward passes. Each pass, due to the randomness of Dropout, produces a slightly different prediction probability distribution. For sample $x_j^u$, we obtain a set of $T$ predictions $\{p_{j,t}^u\}_{t=1}^T$.

We utilize this set of predictions to compute two key metrics:
\begin{enumerate}
    \item \textbf{Mean Prediction}: The average of all predictions is considered the most robust prediction for the sample.
    $$
    \bar{p}_j^u = \frac{1}{T}\sum_{t=1}^{T} p_{j,t}^u
    $$
    From this, we obtain the candidate pseudo-label $\hat{y}_j^u = \text{argmax}(\bar{p}_j^u)$.

    \item \textbf{Prediction Uncertainty}: We use the variance of the $T$ predictions to quantify the model's epistemic uncertainty, i.e., the degree of uncertainty the model has about its own predictions. A commonly used metric is the trace of the covariance matrix of the prediction probabilities:
    $$
    U_j = \text{Tr}(\text{Cov}(\{p_{j,t}^u\}_{t=1}^T))
    $$
    A high variance implies that the model's predictions diverge significantly across different Dropout configurations, indicating unstable judgment for that sample and suggesting its pseudo-label is likely unreliable.
\end{enumerate}

\subsubsection{Principled Pseudo-Label Filtering}
Combining the above two metrics, we propose a principled dual-criterion pseudo-label filtering mechanism. For an unlabeled sample $x_j^u$, its pseudo-label $\hat{y}_j^u$ is adopted for subsequent training only if it satisfies both of the following conditions:
$$
M_j = \mathbb{I}(\max(\bar{p}_j^u) \ge \tau_c \quad \land \quad U_j \le \tau_u)
$$
Here, $\mathbb{I}(\cdot)$ is the indicator function, which is 1 if the condition is true and 0 otherwise. $M_j$ is the validity mask for this sample. $\tau_c$ is a preset confidence threshold, ensuring we only consider predictions where the model is confident; $\tau_u$ is a preset uncertainty threshold, ensuring we only adopt predictions where the model is stable and confident. This strategy effectively filters out high-risk pseudo-labels that arise due to occlusion or class confusion.

\subsection{Overall Loss Function}
The total loss function of RAUM-Net, denoted as $L$, is composed of a supervised loss $L_s$ on labeled data and a consistency loss $L_u$ on unlabeled data, weighted by a hyperparameter $\lambda$:
$$
L = L_s + \lambda L_u
$$
where $\lambda$ is a hyperparameter balancing the two parts of the loss.

The supervised loss $L_s$ is the standard cross-entropy loss calculated on the labeled samples:
$$
L_s = \frac{1}{B_L} \sum_{i=1}^{B_L} H(y_i^l, f_\theta(A_{strong}(x_i^l)))
$$
where $B_L$ is the batch size for labeled samples, and $H(\cdot, \cdot)$ denotes the cross-entropy function.

The unsupervised consistency loss $L_u$ is computed only on unlabeled samples that pass the filtering criteria of our RABU module. It enforces consistency between model predictions on different augmentations (strong augmentation) of the same image and the high-quality pseudo-label:
$$
L_u = \frac{1}{\sum_{j=1}^{B_U} M_j} \sum_{j=1}^{B_U} M_j \cdot H(\hat{y}_j^u, f_\theta(A_{strong}(x_j^u)))
$$
Note that the denominator is the number of samples passing the filter within the batch, $\sum M_j$, which ensures that the loss gradient is adaptively normalized based on the number of valid pseudo-labels in each batch. If no samples pass the filter in a given batch, $L_u$ for that batch is 0. In this manner, RAUM-Net intelligently and safely utilizes unlabeled data, significantly improving learning robustness and final performance in challenging scenarios.

\section{Experiments}
This section details a comprehensive suite of experiments conducted to evaluate our proposed RAUM-Net framework. We begin by introducing the datasets, semi-supervised splits, occlusion simulation strategies, and evaluation metrics employed. Subsequently, we present performance comparison results between RAUM-Net and several state-of-the-art semi-supervised learning methods under standard and occluded conditions. Finally, through in-depth ablation studies, parameter sensitivity analyses, and efficiency evaluations, we systematically validate the effectiveness of each innovative component within RAUM-Net and demonstrate its excellent balance between performance and efficiency.

\subsection{Experimental Setup}

\textbf{Datasets:} We selected two benchmark datasets widely used for evaluating fine-grained visual classification performance: CUB-200-2011 (referred to as CUB) and Stanford Cars (referred to as Cars). The CUB dataset comprises 200 distinct bird species, with the primary challenge being the differentiation of visually similar subspecies. The Stanford Cars dataset contains 196 different car models, posing challenges in recognizing subtle variations due to changes in year, configuration, or viewpoint.

\textbf{Semi-supervised Splits:} To simulate scenarios with high annotation costs and scarce labeled data, as is common in real-world applications, we adopted strict semi-supervised splits for both datasets. Specifically, we randomly sampled 10\% and 50\% of the samples from the training set of each class to serve as labeled data ($X_L$), with the remaining samples designated as unlabeled data ($X_U$). This setup effectively tests the algorithms' learning capabilities under varying degrees of label sparsity.

\textbf{Occlusion Simulation:} Objects in real-world environments are often partially occluded. To systematically study the model's robustness in such situations, we designed a synthetic occlusion generation process. This process involves overlaying one or more gray squares, each covering 20\% (light occlusion) or 40\% (heavy occlusion) of the image area, at random locations within an image. We applied this occlusion to unlabeled training data and all test data to comprehensively assess the model's performance when faced with varying degrees of missing visual information.

\textbf{Evaluation Metric:} Standard Top-1 classification accuracy was used as the primary metric to measure model performance across all experiments.

\textbf{Implementation Framework:} All experiments were performed using the MindSpore deep learning framework.

\subsection{Implementation Details}

\textbf{Baselines:} We compared RAUM-Net against several representative semi-supervised learning methods, including FixMatch \cite{sohn2020fixmatch}, FlexMatch \cite{zhang2021flexmatch}, and AnomalyMatch \cite{gomez2025anomalymatch}, which has shown strong performance in confidence estimation. To ensure a fair comparison, we used VMamba-S, pre-trained on ImageNet-1K, as the backbone for all comparative methods. For data augmentation, all methods employed standard weak augmentation (random horizontal flip and crop) and strong augmentation (RandAugment). We meticulously searched for key hyperparameters (e.g., confidence thresholds) for all baseline methods to achieve their optimal performance.

\textbf{Our Implementation:} Our RAUM-Net is also built upon the VMamba-S backbone. The model was trained end-to-end using the AdamW optimizer with an initial learning rate of $1 \times 10^{-4}$, a weight decay of 0.05, and a cosine annealing schedule for dynamic learning rate adjustment. The batch size was set to 64. In our core RABU module, the number of MC Dropout forward passes ($T$) was set to 10, and the high-confidence filtering threshold ($\tau_c$) was fixed at 0.95. The weight factor for the unsupervised loss ($\lambda$) was set to 1.0, with a warm-up period of 100 epochs for the teacher model. All experiments were conducted in parallel on 4 Ascend 910b NPUs.

\begin{table*}[htbp]
\centering

  \caption{Performance Comparison with SOTA Methods on CUB-200 and Stanford Cars Datasets (Top-1 Accuracy \%)}
  \label{tab:main}
  \begin{tabular}{lcccccc}
    \toprule
    \textbf{Dataset} & \textbf{Method} & \textbf{Label Ratio} & \textbf{No Occlusion} & \textbf{Light Occlusion (20\%)} & \textbf{Heavy Occlusion (40\%)} \\
    \midrule
    \multirow{4}{*}{CUB-200-2011} & FixMatch & 10\% & 16.5 & 12.1 & 8.2 \\
    & FlexMatch & 10\% & 17.2 & 12.8 & 8.9 \\
    & AnomalyMatch & 10\% & 17.8 & 13.5 & 9.6 \\
    & RAUM-Net (Ours) & 10\% & 19.7 & 16.8 & 14.1 \\
    \midrule
    \multirow{4}{*}{CUB-200-2011} & FixMatch & 50\% & 56.8 & 50.1 & 43.5 \\
    & FlexMatch & 50\% & 57.5 & 51.0 & 44.4 \\
    & AnomalyMatch & 50\% & 58.2 & 51.9 & 45.3 \\
    & RAUM-Net (Ours) & 50\% & 60.9 & 57.3 & 52.5 \\
    \midrule
    \multirow{4}{*}{Stanford Cars} & FixMatch & 10\% & 83.9 & 78.2 & 70.4 \\
    & FlexMatch & 10\% & 84.7 & 79.1 & 71.3 \\
    & AnomalyMatch & 10\% & 85.5 & 80.0 & 72.1 \\
    & RAUM-Net (Ours) & 10\% & 87.2 & 84.3 & 79.5 \\
    \midrule
    \multirow{4}{*}{Stanford Cars} & FixMatch & 50\% & 91.1 & 85.4 & 78.2 \\
    & FlexMatch & 50\% & 91.8 & 86.3 & 79.1 \\
    & AnomalyMatch & 50\% & 92.4 & 87.0 & 80.0 \\
    & RAUM-Net (Ours) & 50\% & 94.0 & 91.2 & 86.8 \\
    \bottomrule
  \end{tabular}
\end{table*}

\subsection{Main Results and Analysis}
Table \ref{tab:main} summarizes the detailed performance comparisons between RAUM-Net and the baseline methods across both datasets, two label ratios, and three occlusion conditions.

The experimental results clearly demonstrate that RAUM-Net exhibits consistent and significant superiority across all tested conditions. In the CUB 10\% setting with extremely sparse labels, baseline performance in the no-occlusion scenario hovered between 16\%-18\%, fully reflecting the task's difficulty. Under these stringent conditions, RAUM-Net still achieved an accuracy of 19.7\%, showcasing its strong foundational learning capability. When the label count increased to 50\%, all methods showed substantial performance improvements, with RAUM-Net consistently maintaining a lead of approximately 2-3 percentage points.

RAUM-Net's notable ability is its superior robustness against occlusion. As observed in Table \ref{tab:main}, a clear trend emerges: with increasing occlusion, all methods experienced a sharp performance decline, yet RAUM-Net's performance advantage continued to widen. Under the most challenging heavy occlusion conditions, for the CUB dataset with 50\% label ratio and heavy occlusion, RAUM-Net's accuracy (52.5\%) was 7.2\% higher than AnomalyMatch (45.3\%), the next best performing method. This phenomenon profoundly reveals the vulnerability of traditional semi-supervised methods when faced with incomplete visual information. Methods like FixMatch rely on a single confidence measure from the model's output. When critical features are occluded, the model is highly prone to producing erroneously confident predictions. These toxic pseudo-labels propagate and accumulate during training, ultimately leading to model performance collapse. In contrast, RAUM-Net effectively avoids this pitfall through its dual-safety mechanism. On one hand, the Regional Attention (RA) component intelligently guides the model to disregard occluded regions and instead extract key discriminative information from the remaining, unobscured areas, such as a bird's head or a car's wheel hub. On the other hand, the Bayesian Uncertainty (BU) component acts as a quality inspector, accurately identifying samples whose model predictions are unstable due to occlusion and filtering them out based on a principled uncertainty metric, thereby ensuring that pseudo-labels used for guiding learning maintain high purity. It is this design that enables RAUM-Net to perform robust and efficient semi-supervised learning even in demanding environments.

\subsection{Ablation Study}
To further dissect RAUM-Net's internal workings and the contributions of each component, we conducted detailed ablation experiments and related parameter sensitivity studies.

\begin{table*}[htbp]
  \centering
\caption{Ablation Study of RAUM-Net Components on CUB-200 (10\% Label, 40\% Heavy Occlusion)}
\label{tab:ablation}
  \begin{tabular}{lccccc}
    \toprule
    \textbf{Model Configuration} & \textbf{RA} & \textbf{BU} & \textbf{Backbone} & \textbf{Accuracy (\%)} & \textbf{Performance Change} \\
    \midrule
    Baseline (FixMatch) & suiek & None & ResNet-50 & 4.5 & - \\
    Baseline + Mamba & None & None & VMamba-S & 8.2 & +3.7 vs \#1 \\
    RAUM-Net w/o RA & None & Contains & VMamba-S & 11.5 & +3.3 vs \#2 \\
    RAUM-Net w/o BU & Contains & None & VMamba-S & 10.6 & +2.4 vs \#2 \\
    RAUM-Net (Complete Model) & Contains & Contains & VMamba-S & 14.1 & +5.9 vs \#2 \\
    \bottomrule
  \end{tabular}

\end{table*}

We performed a component-wise analysis of RAUM-Net on the CUB-200 dataset with 10\% labels and heavy occlusion, the most challenging setting, with results presented in Table \ref{tab:ablation}. Replacing the backbone from ResNet-50 to VMamba-S (\#2 vs \#1) yielded a significant 3.7 percentage point improvement, eloquently demonstrating the inherent advantages of the Mamba architecture in modeling long-range dependencies and capturing fine-grained local details, making it a superior choice for FGVC tasks. Furthermore, introducing the Bayesian uncertainty module alone (\#3) or the regional attention module alone (\#4) both led to substantial performance gains over the Mamba baseline (3.3\% and 2.4\% respectively), clearly quantifying the independent contribution of each component. The most critical finding is that the full RAUM-Net (\#5) achieved a performance gain over the Mamba baseline that exceeded the sum of the individual gains from RA and BU modules. This indicates a positive synergistic effect between regional attention and Bayesian uncertainty: regional attention provides a cleaner and more discriminative feature input for uncertainty estimation, leading to more accurate uncertainty judgments; and the more reliable pseudo-label filtering mechanism, in turn, helps the regional attention module better learn which regions are truly worth focusing on. It is this virtuous cycle that enables RAUM-Net as a whole to exhibit power far beyond the simple summation of its parts.

\begin{figure}[!ht]
\centering
\includegraphics[width=\columnwidth]{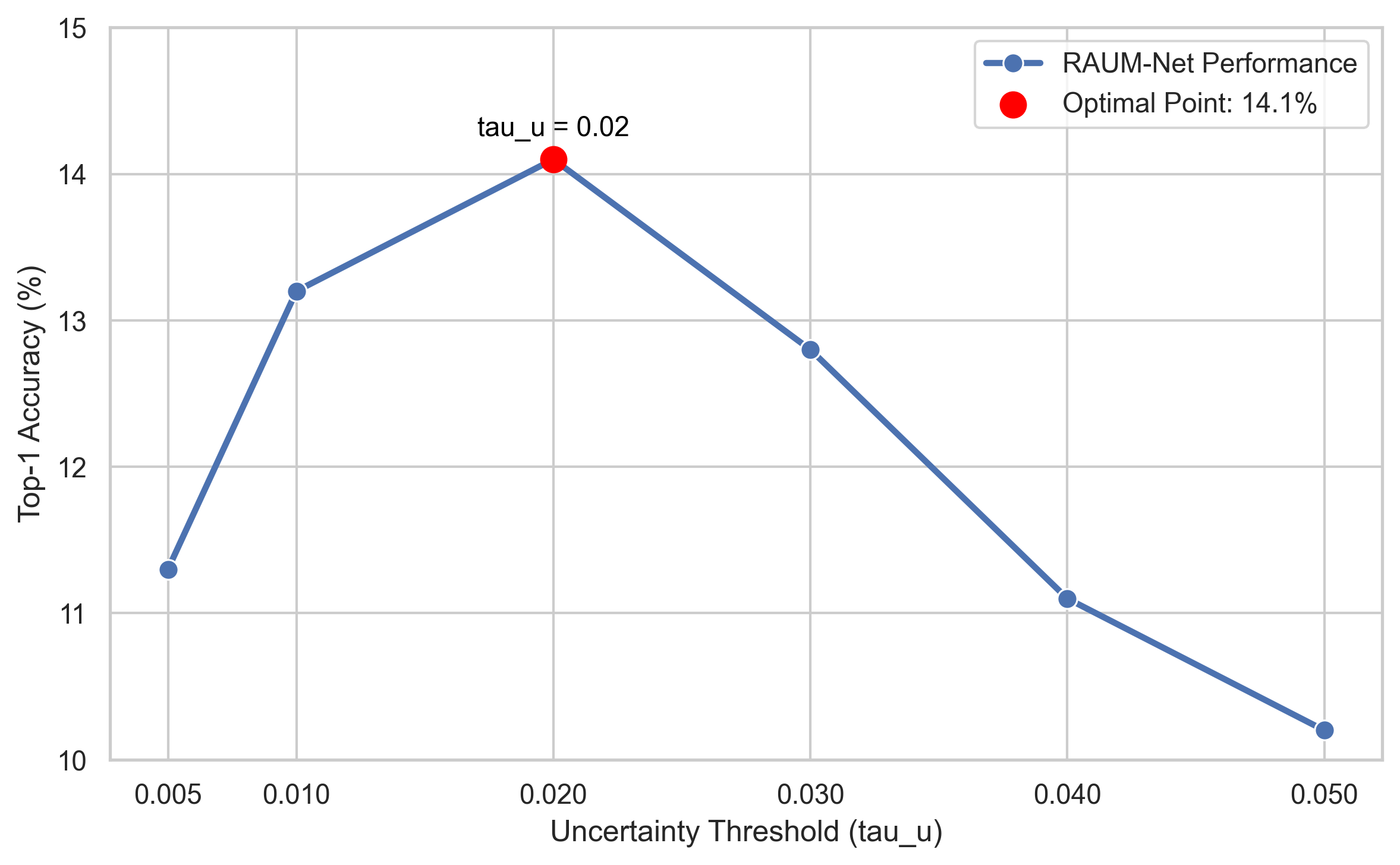}
    \caption{Sensitivity analysis of the uncertainty threshold ($\tau_u$). The plot shows the Top-1 Accuracy (\%) of RAUM-Net across different values of $\tau_u$ on the CUB-200 dataset with 10\% labels and 40\% heavy occlusion. }
\label{fig:sensitivity}
\end{figure}

\textbf{Sensitivity Analysis:} We also investigated the impact of a key hyperparameter in the RABU module, the uncertainty threshold $\tau_u$, on model performance. As illustrated in Figure \ref{fig:sensitivity}, the choice of $\tau_u$ is crucial for final performance. An overly low $\tau_u$ results in excessively strict pseudo-label filtering, wasting a large number of valuable unlabeled samples and leading to insufficient model learning. Conversely, an overly high $\tau_u$ relaxes the filtering criteria, introducing too much noisy labeling and compromising training stability. Experiments reveal that an optimal range for $\tau_u$ exists, which balances the quality and quantity of pseudo-labels to achieve the best classification performance. This also provides a basis for exploring adaptive threshold adjustment strategies in future work.

\subsection{Model Efficiency Evaluation}
In addition to pursuing superior classification accuracy, we also focused on the practicality and efficiency of the model. In Table \ref{tab:efficiency}, we compare RAUM-Net with several baseline models in terms of model parameters (Params), computational complexity (FLOPs), and inference speed (FPS) on a single Ascend 910b NPU.

\begin{table}[htbp]
  \centering
\caption{Model Efficiency Comparison. A summary of model parameters, computational complexity, and inference speed. Inference speed (FPS) was measured on a single Ascend 910b.}
\label{tab:efficiency}
  \begin{small}
  \begin{tabular}{lccc}
    \toprule
    \textbf{Model} & \textbf{Params (M)} & \textbf{FLOPs (G)} & \textbf{FPS} \\
    \midrule
    FixMatch + ResNet-50 & 25.6 & 4.1 & 78 \\
    FixMatch + VMamba-S & 26.3 & 4.5 & 72 \\
    RAUM-Net (Ours) & 26.4 & 4.9 & 65\\
    \bottomrule
  \end{tabular}
  \end{small}
\end{table}

As shown in Table \ref{tab:efficiency}, a direct comparison of RAUM-Net with baseline models across three key efficiency metrics is presented: model parameters (Params), computational complexity (FLOPs), and inference speed (FPS). RAUM-Net demonstrates excellent performance in terms of model efficiency. Its core RABU module is designed to be extremely lightweight, introducing only a negligible increase in parameters (approximately 0.1M) and computational overhead (approximately 0.4G FLOPs) compared to the base VMamba-S backbone. Consequently, RAUM-Net's inference speed is comparable to baseline methods using the same VMamba-S backbone. This indicates that our method provides performance improvements in occluded scenarios without a significant additional computational burden, proving it to be an efficient and practical solution with great potential for deployment in resource-constrained real-world applications.

\section{Conclusion}
This paper addresses the challenges in semi-supervised fine-grained visual classification (FGVC) under occlusion and label scarcity by proposing a novel framework named RAUM-Net. This method combines the efficient Mamba backbone with our innovative Regional Attention and Bayesian Uncertainty (RABU) module. The Mamba architecture provides a powerful foundation for extracting fine-grained features, while the RABU module focuses on unobscured key regions through its regional attention mechanism and enforces strict pseudo-label filtering via Bayesian uncertainty quantification. Together, these components synergistically tackle the dual difficulties of feature ambiguity and pseudo-label noise. Extensive experiments conducted on multiple FGVC benchmark datasets demonstrate that RAUM-Net significantly outperforms existing state-of-the-art methods across various semi-supervised and occluded conditions. Notably, under heavy occlusion or extreme label sparsity, our method exhibits remarkable robustness and higher classification accuracy, proving its substantial potential for addressing challenging real-world problems. Future work will explore adaptive attention and uncertainty threshold strategies and extend the ideas of this framework to other fine-grained visual tasks. We believe that RAUM-Net offers an effective and practical solution for the field of semi-supervised FGVC.

\section{Acknowledgments}
We extend our sincere gratitude to the MindSpore community for providing a robust and efficient deep learning framework that facilitated our research. We are also thankful to the OpenI platform for offering valuable computing resources and support, which were instrumental in conducting the extensive experiments presented in this paper.

{
    \small
    \bibliographystyle{ieeenat_fullname}
    \bibliography{main}
}

\end{document}